\documentclass[letterpaper, 10 pt, conference]{./support/ieeeconf}
\usepackage{times}
\usepackage[pdftex]{graphicx}
\usepackage{subfigure}
\usepackage{amsmath,amssymb,amsopn,amstext,amsfonts}
\usepackage{cancel}
\usepackage[space]{cite}
\usepackage{pdfsync}
\usepackage{balance}
\usepackage{color}
\usepackage{mathtools}
\usepackage[ruled,vlined]{algorithm2e}
\usepackage{bm}

\usepackage{diagbox}
\usepackage{float}
\usepackage{epstopdf}
\usepackage{pifont}
\usepackage{fixltx2e}
\usepackage{amsmath}
\usepackage{multirow}
\usepackage{url}
\usepackage{svg}
\usepackage{verbatim}
\usepackage[linkcolor=black,citecolor=black,urlcolor=black,colorlinks=true]{hyperref}
\graphicspath{{./figures/}}
\DeclareGraphicsExtensions{.png,.jpg,.eps,.pdf}
\IEEEoverridecommandlockouts
\title{\LARGE \bf Autonomous Flights in Dynamic Environments with Onboard Vision }
\author{Yingjian Wang, Jialin Ji, Qianhao Wang, Chao Xu and Fei Gao
\thanks{
	All authors are with the State Key Laboratory of Industrial Control Technology, Institute of Cyber-Systems and Control, Zhejiang University, Hangzhou, 310027, China, and also with the Huzhou Institute of Zhejiang University, Huzhou, 313000, China.
    \tt\small \{yj\_wang, jlji, qhwangaa, cxu, fgaoaa\}@zju.edu.cn}
}
\begin{document}

\maketitle
\thispagestyle{empty}
\pagestyle{empty}

\begin{abstract}
In this paper, we introduce a complete system for autonomous flight of quadrotors in dynamic environments with onboard sensing. Extended from existing work, we develop an occlusion-aware dynamic perception method based on depth images, which classifies obstacles as dynamic and static. For representing generic dynamic environment, we model dynamic objects with moving ellipsoids and fuse static ones into an occupancy grid map. To achieve dynamic avoidance, we design a planning method composed of modified kinodynamic path searching and gradient-based optimization. The method leverages manually constructed gradients without maintaining a signed distance field (SDF), making the planning procedure finished in milliseconds. We integrate the above methods into a customized quadrotor system and thoroughly test it in real-world experiments, verifying its effective collision avoidance in dynamic environments.
    
\end{abstract}

\section{Introduction}
\label{sec:introduction}
Thanks to the maturity of autonomous navigation technology, aerial robots, especially quadrotors, have made impressive progress in the last decade. However, it is still challenging to let quadrotors operate autonomously in dynamic environments due to trifold reasons. Firstly, the perception of dynamic obstacles is hard to satisfy the efficiency and accuracy requirements at the same time, making the quadrotor fragile in a complex dynamic environment. Moreover, to make a quadrotor fly smoothly among moving obstacles, the onboard planner must fully utilize the obstacle position and velocity estimation while retaining high efficiency. Finally, the system integration requires robustness from all modules, including localization, perception, and planning, under limited onboard sensing and computing resources. These challenges render autonomous flights in an unknown dynamic environment hard, making there always lacks a convincing systematic solution in the community.

In this paper, we bridge this research gap by introducing a complete system composed of an accurate dynamic environment perception module and an accompanied trajectory generation method. For dynamic environment perception, we classify the obstacles into static ones and dynamic ones by clustering, tracking, and voting the point cloud data from a depth sensor, inspired by \cite{eppenberger2020leveraging}. To address the issue of occlusion, we propose an occlusion-aware Kalman filter to estimate the motion states of dynamic objects, including position, velocity, size, and associated uncertainty, making the classification more accurate and robust. We then fuse the motionless points into an occupancy grid map and model the dynamic ones as several moving ellipsoids for the subsequent trajectory generation.

Considering the static and the dynamic obstacles simultaneously, we propose a hierarchical trajectory generation method. Firstly, we design a kinodynamic path searching method to search for a safe, feasible, and minimum-time initial path by dynamic safety check. After parameterizing the initial path into a B-spline curve, we refine the curve with a carefully designed gradient-based optimization. Extending our previous work \cite{zhou2020ego}, we obtain gradients to avoid the static and dynamic obstacles by manually constructed penalties without maintaining an SDF, which dramatically lowers the computational burden.

\begin{figure}[t]
	\begin{center}
		\includegraphics[width=1.0\columnwidth]{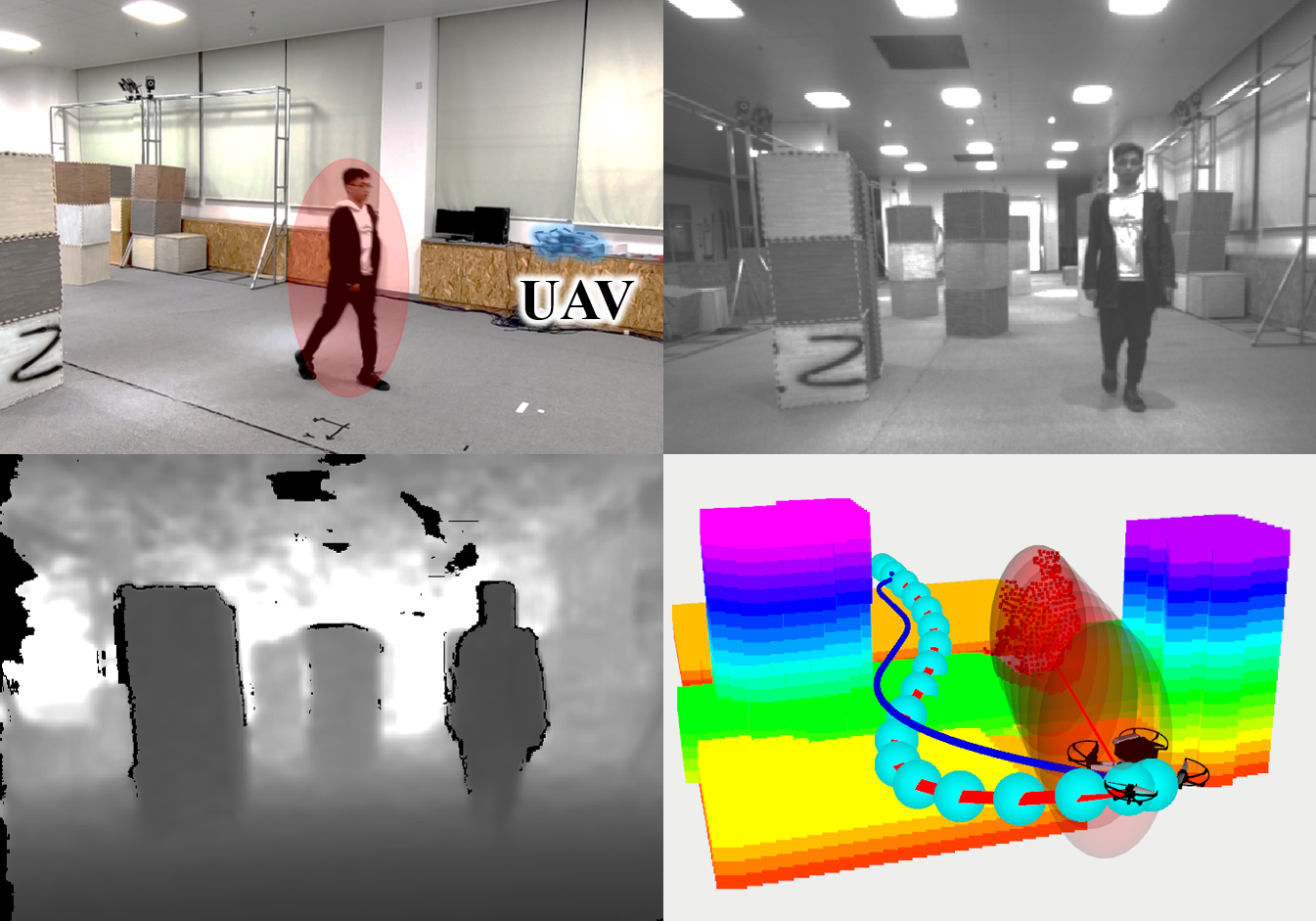}
	\end{center}
	\caption{
		\label{fig:map} Experimental results of autonomous flight in a dynamic environment with onboard sensing. Top left: a flying UAV in front of a walking person. Top right: onboard grayscale image. Bottom left: depth image. Bottom right: the optimized B-spline trajectory (blue points and red curve) between the detected moving person(red ellipsoids with a red line indicating the velocity) and the static grids.
	}
	\vspace{-0.3cm}
\end{figure}

Compared to existing state-of-the-art works\cite{lin2020robust,zhou2020ego}, our proposed method can generate safe trajectories in more generic environments composed of cluttered static and dynamic obstacles. We perform comprehensive tests in simulation and real world to validate the robustness of our method. Summarizing our contributions as follows:
\begin{itemize}
	\item[1)] We develop an occlusion-aware dynamic environment perception method based on \cite{eppenberger2020leveraging}, which efficiently yet accurately classifies the obstacles into static ones and dynamic ones. Moreover, we represent both classes of obstacles as two different data structures to handle generic complex environments.
	\item[2)] We propose a hierarchical trajectory generation method composed of a kinodynamic searching method and a gradient-based optimization. We leverage the environment representation mentioned above to construct gradients manually, which avoids the overhead of maintaining an SDF, making the whole procedure finished in milliseconds.
	\item[3)] We integrate our proposed methods into a fully autonomous quadrotor system and release our software for the community's reference.
\end{itemize}

\section{Related Work}
\label{sec:related_work}
\subsection{Dynamic Perception}
There are mainly two categories of methods for dynamic perception. One is to use imagery data. In \cite{cheng2017autonomous}, authors adopt the Frame-Difference (FD) method \cite{zhan2007improved} to detect moving objects, which requires that UAVs hover steady. However, frequent hovering will greatly affect the smoothness of flight in dynamic avoidance missions. Besides, some work utilize outliers of the feature tracking module in feature-based vision system to detect dynamic objects\cite{xu2019mid, berker2017feature, barsan2018robust}. However, it requires dense feature points, which is hard to satisfy in generic environments. Extended from optical flow, scene flow method \cite{wu2019pointpwc, francis2015detection} computes 3D velocity of each pixel. This technique, however, is unable to run in real-time on a computationally-constrained platform. Another approach is using detector \cite{redmon2018yolov3}, or segmentation networks \cite{he2017mask}, which behaves well in detection of predefined classes such as pedestrians or cars \cite{zhang2017towards, jafari2014real}, but cannot handle generic cases.

Apart from using imagery data, point-cloud-based methods aim to estimate the motion of objects leveraging 3D information. Kinect Fusion \cite{newcombe2011kinectfusion} implements Iterative Closest Point (ICP)\cite{bouaziz2013sparse} to deal with moving objects. However, this approach would result in high computational overhead. Some work \cite{oleynikova2015reactive, skulimowski2019interactive} use U-map \cite{arnell2005fast} to detect obstacles which considers that points with similar depth belong to the same object. This method divides the point cloud into individual objects without dynamic classification, which does not represent the dynamic environment accurately. Our approach builds upon and extends a dynamic obstacle detection and tracking algorithm \cite{eppenberger2020leveraging}, which tracks clusters and classifies them as dynamic or static. Varying from them, we conduct tracking with occlusion-awareness and replace their human detector with our proposed re-free strategy to address the issue of erroneous occupancy caused by temporary standstills.

\subsection{Dyanmic Planning}

Most planning methods in dynamic environments inherit the ones in static environments and just design some techniques to handle the moving obstacles.
Reactive methods, including velocity obstacles (VO) \cite{fiorini1998motion, wilkie2009generalized, van2011reciprocal}, inevitable collision states (ICS) \cite{fraichard2004inevitable, althoff2012safety} and artificial potential field (APF) \cite{malone2017hybrid}, leverage the information of current moving obstacles and only compute one-step actions, which makes them quite short-sighted due to the lack of long horizon. Seder et.al  \cite{seder2007dynamic} first adopt a focused D* to find a reference path without considering obstacle motions and then generates admissible local trajectories around the reference path using a dynamic window algorithm. In \cite{chiang2017dynamic}, the RRT and forward SR sets are utilized to find the reference path and avoid moving obstacles, respectively.

In recent studies, Gao et.al\cite{gao2017quadrotor} use semi-definite relaxation on nonconvex Quadratical Constraint Quadratic Programming (QCQP) to optimize a piecewise polynomial trajectory in dynamic environments. However, this method and other global planners \cite{cao2019dynamic, zhu2020online}, assume that locations and velocities of all obstacles are known, which cannot work in autonomous drone platform. In \cite{lin2020robust}, the authors obtain objects' necessary state information by onboard detection method and utilize chance-constrained nonlinear model predictive control for obstacle avoidance in dynamic environments. However, it regards all obstacles as ellipsoids, which makes it cannot handle complex static obstacles. Beyond that, nonlinear optimization is prone to fall into a local minimum without a proper initial guess.

\section{Dynamic Environment Perception}
\label{sec:dynamic environment perception}

In this section, we describe a dynamic perception module built upon \cite{eppenberger2020leveraging}. Given a point cloud generated from depth image at time $\tau \in \mathbb{R}$, we filter it to reduce sensor noise and lower computational overhead, and then cluster the filtered point cloud into individual objects per frame using DBSCAN\cite{DBSCAN}, resulting in a set of $\emph{m}$ clusters $\emph{C}_\tau =  \{\emph{c}^{1}_\tau, \emph{c}^2_\tau, ..., \emph{c}^m_\tau \}$. To estimate moving agents' necessary states, we then conduct occlusion-aware tracking and dynamic classification.

\begin{figure*}[h]
	\begin{center}
		\includegraphics[width=2.0\columnwidth]{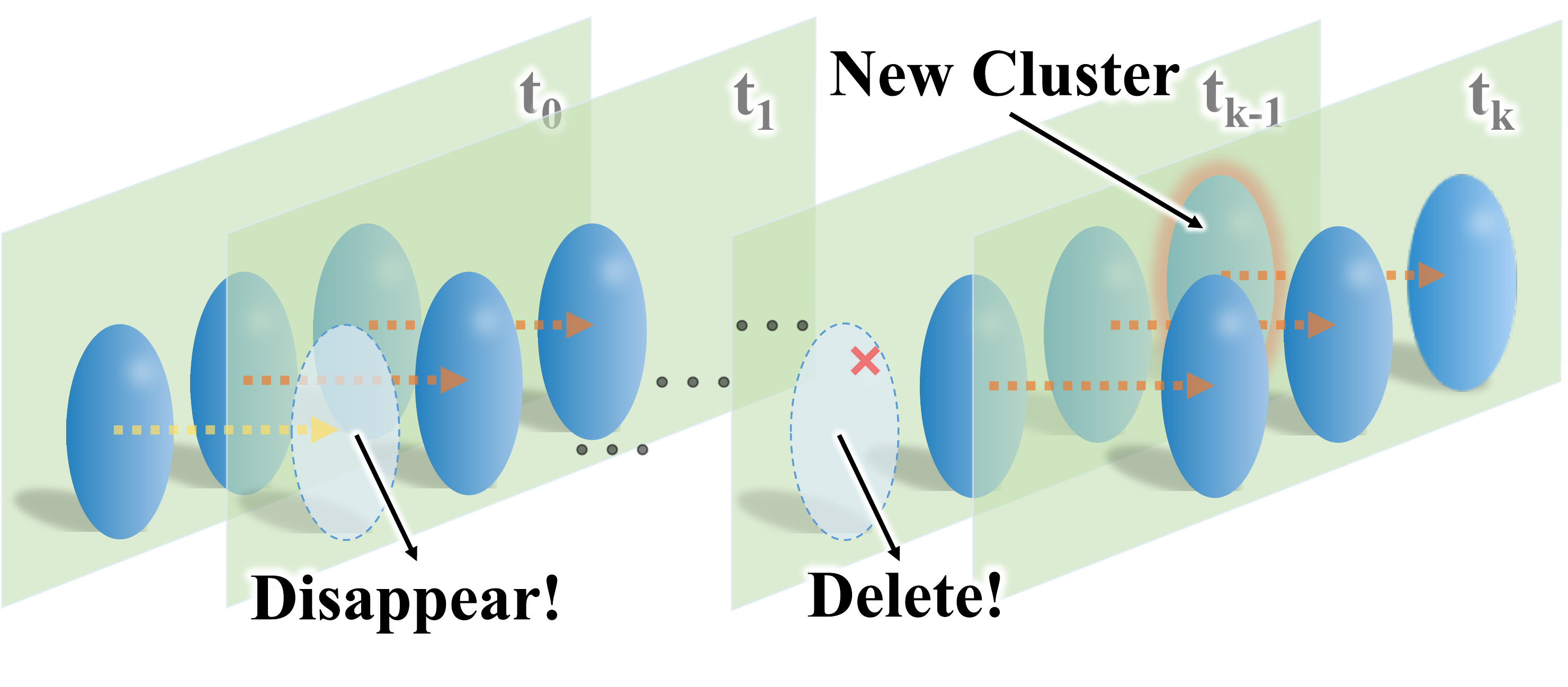}
	\end{center}
	\caption{
		\label{fig:track}
		Non-connected KFs in the previous frame will keep propagating until their corresponding cluster disappears for specific frames (i.e., the white ellipsoids in t$_1$ and t$_{k-1}$), while non-connected clusters in the current frame will be marked as new objects and begin their tracking (frame t$_{k-1}$).
	}
	\vspace{0.0cm}
\end{figure*}

\subsection{Occlusion-aware Tracking}
\label{sec:Occlusion-aware Tracking}

We denote a list of cluster sets over \emph{k} frames separated by $\Delta\emph{t}$ as $\widetilde{\emph{C}}_{t,k}$ = \{\emph{C}$_{t-\emph{k}\cdot\Delta\emph{t}}$,..., \emph{C}$_{t-\Delta\emph{t}}$, \emph{C}$_{t}$\}. Given $\widetilde{\emph{C}}_{t,k}$, we use an occlusion-aware tracking algorithm extended from \cite{eppenberger2020leveraging} to obtain their $\textbf{tracking history}$ $\emph{H}_{t}^{ i} = \{\emph{c}_{t-\emph{n}\cdot\Delta\emph{t}}^{*},..., \emph{c}_{t-\Delta\emph{t}}^{*}, \emph{c}_{t}^{i}\}$ which include indices of the same cluster.

\begin{figure}[h]
	\begin{center}
		\includegraphics[width=1.0\columnwidth]{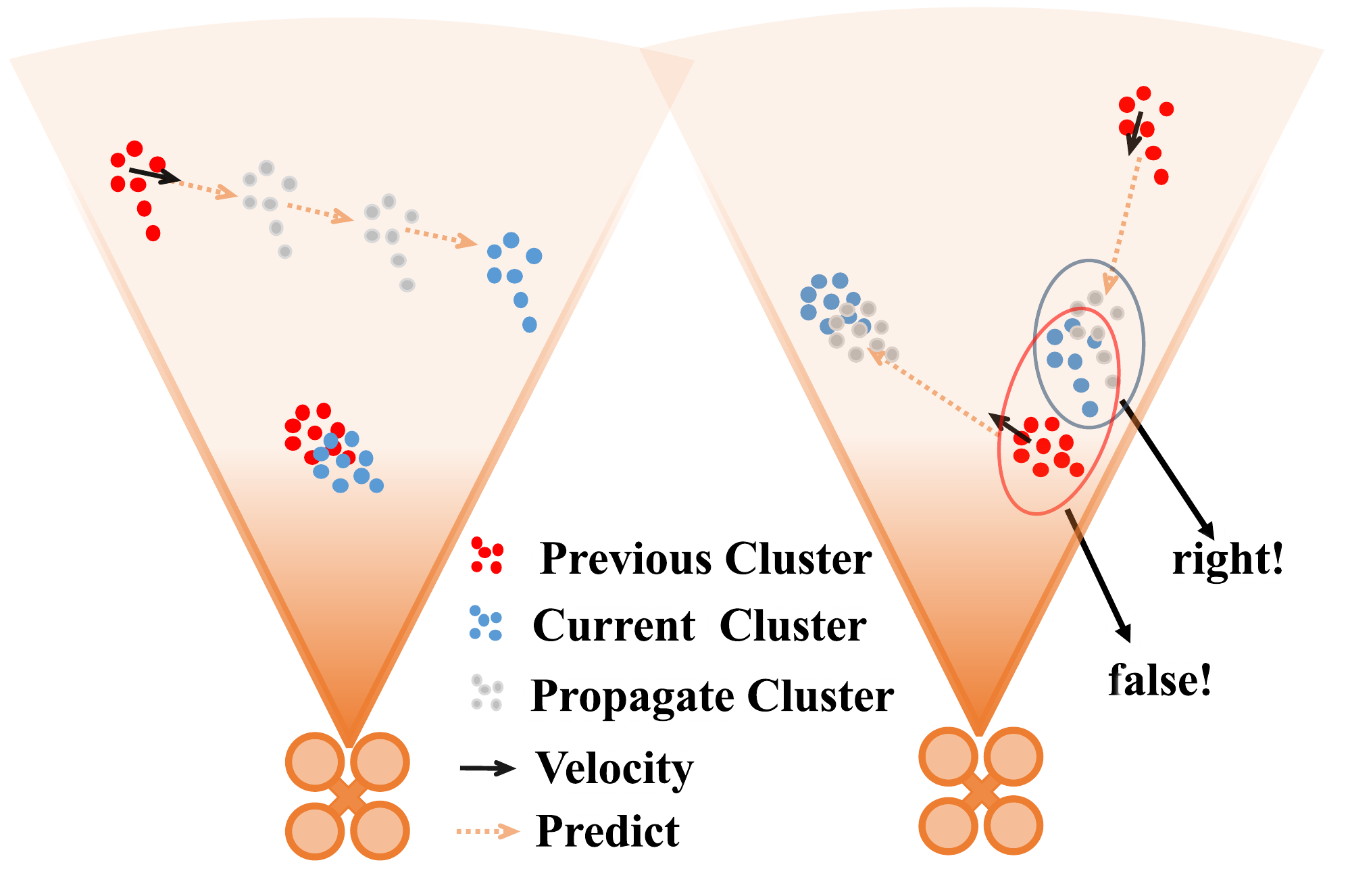}
	\end{center}
	\caption{
		\label{fig:cluster}
		The left figure shows our tracking algorithm could address the issue of occlusion. The right figure shows our method could avoid incorrect association, which would happen in the association without propagation.
	}
	\vspace{-0.5cm}
\end{figure}

As shown in Fig.\ref{fig:track}, for each cluster at time $t_0$, we will apply a Kalman filter to it initially with a conservative motion model,  similar to the work of Azim et al.\cite{azim2012detection}, resulting in a list of Kalman filters $\emph{K}_{t_0} =  \{\kappa^{1}_{t_0}, \kappa^2_{t_0}, ..., \kappa^l_{t_0} \}$. Then, in every later frame at time $t_i$, we firstly propagate all Kalman filters in $\emph{K}_{t_{i-1}}$ forward and then associate centroids of clusters in the current frame to the forward propagated positions by k-Nearest Neighbor searching \cite{tian2017knn}. If the association of cluster \emph{c}$_{t_0}^{j}$ successes, the cluster will inherit its corresponding tracking history. Otherwise, we suppose the unassociated cluster is a newly appeared object and create a new Kalman filter for it. To address the short-time occlusion issue, we mark the unassociated clusters lost and keep their Kalman filters running until they are tracked or lost for a threshold time $t_d$, as shown in the right figure of Fig.\ref{fig:cluster}.

Moreover, it is essential that we attempt to associate current clusters to forward propagated Kalman filters rather than clusters in the previous frame like \cite{eppenberger2020leveraging}. As shown in the right figure of Fig.\ref{fig:cluster}, when there are several moving agents, the association without considering motion information may be incorrect due to the movements of dynamic obstacles. Compared with it, we track clusters with forward propagated Kalman filters, efficiently avoiding plenty of incorrect associations.

\subsection{Classification as Dynamic or Static}
\label{sec:Classification as Dynamic or Static}

We then classify the generated cluster as either static or dynamic. Inspired by \cite{eppenberger2020leveraging}, we obtain the motion of clusters by comparing point clouds from two frames being $\delta$ seconds apart. Firstly, we build kdTree with the dense, non-filtered point cloud from the  reference frame at time $t-\delta$. Then we measure the global nearest neighbor distance \emph{d}$^{i,j}$ of each point p$^{i,j}$ belonging to the cluster \emph{c}$_{t}^{i}$ in the current frame at time $t$ by the kdTree built before.  Then velocity of each point is calculated by $\emph{v} = \frac{\emph{d}^{i,j}}{\delta}$ and the points with $v >  v_{d}$ vote for dynamic. After voting, a cluster will be considered dynamic obstacle if the absolute or relative amount of votes for being dynamic surpass respective thresholds $l_{dyn}^{abs}$ or $l_{dyn}^{rel}$.

To guarantee that each point must be observed in both current frame and reference frame,  there are two cases that should be noted to improve performance of classification. One is that when the FOV changes between two frames, we only allow points which appear in the overlapped area to vote. Another is the occlusion caused by dynamic objects. To address the issue, firstly we associate current cluster \emph{c}$_{t}^{i}$ to its corresponding cluster \emph{c}$_{t-\delta}^{j}$ by the tracking history $\emph{H}_{t}$ and project their 3D centroids onto the image plane. Self-occlusion is supposed to happen if \emph{depth}[\emph{q}$_{t-\delta}^{j}$] $<$ \emph{depth}[\emph{q}$_t^i$] and all points of \emph{c}$_{t}^{i}$ can vote, where \emph{q} denotes the projected 2D point of 3D point \emph{p}. Otherwise, for points \emph{p}$_t$ $\in \mathbb{R}^{3}$ belonging to other clusters at time \emph{t}, we firstly reproject them onto the image plane at time \emph{t-$\delta$},

\begin{align}
	\begin{bmatrix}
		\hat{\emph{q}}_{t-\delta} \\
		1\\
	\end{bmatrix} = \emph{T}_{t-\delta}^{-1} 
	\begin{bmatrix}
	\emph{p}_{t} \\
	1\\
	\end{bmatrix},
\end{align}

\begin{align}
	\begin{bmatrix}
	\emph{q}_{t-\delta}\\
	1\\
	\end{bmatrix}
	 = \emph{K} 
	\hat{\emph{q}}_{t-\delta}, 
\end{align}

where \emph{T}$_{t-\delta}$ is the transformation matrix of camera pose at time $t-\delta$, \emph{K} is the intrinsic matrix of the camera and $\hat{\emph{q}}_{t-\delta}$ is 3D point in the camera frame.

We then suppose point \emph{p}$_t$ is occluded if $|\hat{\emph{q}}_{t-\delta}|_Z - \emph{depth}[\emph{q}_{t-\delta}] > \epsilon$, where $|\bullet|_Z$ denotes the z-coordinate of a 3D point and  the \emph{depth}[\emph{q}$_{t-\delta}$] and $\epsilon$ is set related to the velocity threshold $v_{d}$ to reduce sensor noise. Occluded points will be excluded from vote.

Finally, to improve the robustness of perception, it is essential to check the consistency of classification results over a time horizon $t_h$. Namely, a cluster is judged as dynamic or static finally only if its classification results in all frames over $t_h$ keep the same. 

\subsection{Dynamic Environment Representation}
\label{section:Dynamic Environment Representation}

In order to allow for autonomous flight of UAV in a dynamic environment, we represent the obstacles respectively according to their dynamic or not.

For dynamic agents, we model them as moving 3D ellipsoids with velocity $\textbf{v}^o$:

\begin{align}
	(\textbf{p}-\textbf{p}^o)^T\Theta^o(\textbf{p}-\textbf{p}^o) = 1,
\end{align}

where $\Theta^o = R^Tdiag(\frac{1}{(l_x+r)^2}),\frac{1}{(l_y+r)^2},\frac{1}{(l_z+r)^2})R$. $\textbf{p}^o, l_x, l_y, l_z, r$ are estimated position and axis-length of ellipsoid and inflate radius.

For classified static point cloud, we fuse it into occupancy 3D grid map, which is efficient and easy to use for trajectory planning. Significantly, we introduce \textbf{Re-Free} strategy to eliminate the erroneously occupied space caused by the temporary standstill of moving objects and address the issue of time-lag in \cite{eppenberger2020leveraging} caused by $\delta$ in classification. In detail, we fuse all points in current frame into the occupancy grid map initially.  Furthermore, only if an agent is considered dynamic, we free all related occupancy grids according to its tracking history. In this way, the quadrotor may treat moving objects as static obstacles in the beginning and re-frame them after classification. Sufficient experiments prove that this implementation could significantly improve the robustness and safety of UAVs with limited FOV in obstacle avoidance missions. We finally represent the dynamic environment with moving ellipsoids and occupancy grids.
  
\section{Dynamic Planning}
\label{dynamic planning}

\begin{figure}[t]
	\begin{center}
		\includegraphics[width=1.0\columnwidth]{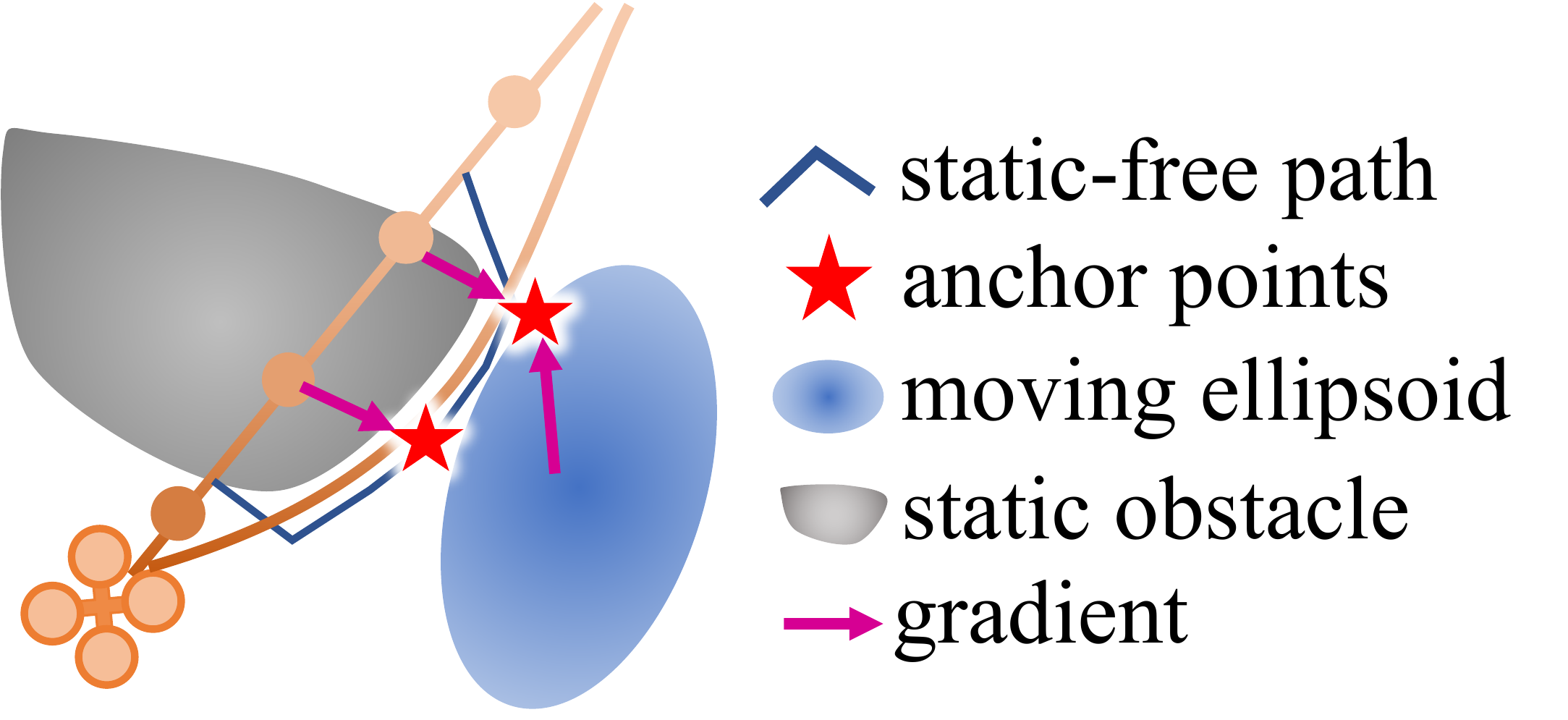}
	\end{center}
	\caption{
		\label{fig:localmin}
		Our previous work\cite{zhou2020ego} search a collision-free path around static obstacles to generate gradients between control points and anchor points. However, these gradients may conflict with gradients from dynamic obstacle and make trajectory into local minimum. Thus, a path searching method considering dynamic objects is required.
	}
	\vspace{-0.3cm}
\end{figure}

In this section, we describe an ESDF-free gradient-based dynamic planning method based on our previous work\cite{zhou2020ego}. In \cite{zhou2020ego}, anchor points are constructed manually to generate gradients by searching a collision-free path around each colliding segment of the initial trajectory. However, this procedure cannot handle dynamic obstacles, which makes it get stuck into a local minimum easily, shown in Fig.\ref{fig:localmin}. In fact, most gradient-based planning methods like \cite{lin2020robust} would face the same problem if the initial trajectory clamps between two near obstacles. 

To solve this problem, we obtain a dynamic obstacles-avoiding initial trajectory for subsequent optimization by a dynamic obstacle avoiding kinodynamic searching method.

\subsection{Dynamic Obstacles-avoiding Kinodynamic Searching}
\label{Dynamic Obstacles-avoiding Kinodynamic Searching}

Similar to \cite{zhou2019robust}, our method generates motion primitives by sampling the control inputs $u_i$ and durations $T_i$ for each node. We define the cost of a trajectory as

\begin{align}
	J(T) = \int_{0}^{T} || u(\tau) ||^2 d\tau + \rho T, 
\end{align}
where $T$ is the total duration and $\rho$ is a tunable weight. Heuristic cost is defined as $J(T_h)$, where $T_h$ is obtained by solving a closed form Optimal Boundary Value Problem.

\begin{figure}[t]
	\begin{center}
		\includegraphics[width=1.0\columnwidth]{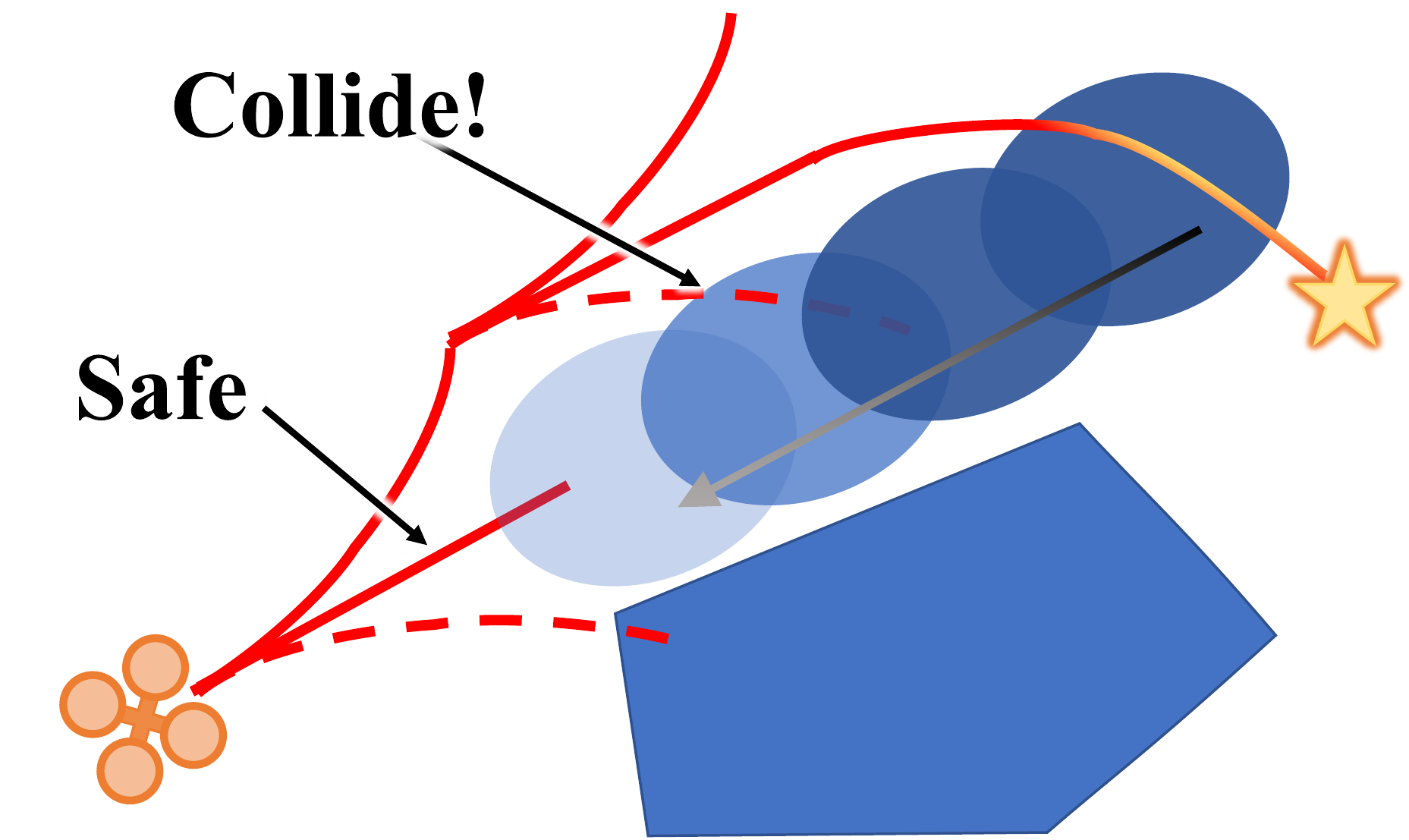}
	\end{center}
	\caption{
		\label{fig:Astar}An illustration of the dynamic obstacle-avoiding kinodynamic path searcher. The red curve indicates the motion primitives (full line means safe and dotted line means unsafe), while the yellow curve represents the analytic expansion scheme which speeds up searching \cite{zhou2019robust}.  Importantly, we predict the motion of dynamic obstacle and check that if the state at moment $\tau$ collides with predicted ellipsoid at the same time.  
	}
	\vspace{-0.5cm}
\end{figure}

Differently, the valid check of each motion primitive is modified by considering the collision check of the dynamic obstacles represented as ellipsoids, as shown in Fig.\ref{fig:Astar}. If the collision condition between the motion primitive at time $\tau$ and the predicted ellipsoid at the same time is satisfied, this motion primitive will be considered unsafe. The minimum distance between a point and an ellipsoid can not perform in closed form \cite{uteshev2018point}. Thus, the collision condition is
\begin{align}
	C_{ij} := \{||\mathbf{p_i} - \mathbf{p_j^o}||_{\Theta^o} \leq 1\},
\end{align}

where $\mathbf{p_j^o}$ and $\Theta^o$ are the estimated position and ellipsoid matrix in  \ref{section:Dynamic Environment Representation}. After searching for an initial trajectory that lies near the optimum, our subsequent B-spline based optimization will find that optimum.

\subsection{B-Spline Trajectory Optimization}

Based on our previous work\cite{zhou2020ego}, the trajectory is parameterized by a uniform B-spline curve $\Phi$, which is a piecewise polynomial uniquely determined by its degree \emph{p}$_b$, a knot span $\Delta t$, and \emph{N}$_c$ control points $Q_{i} \in \mathbb{R}^3$. 

According to the property that the \emph{k}$^{th}$ derivative of a B-spline is still a B-spline with degree $\emph{p}_{p,k} = \emph{p}_b - k$, the velocity, acceleration and jerk control points $\textbf V_i$, $\textbf A_i$ and $\textbf J_i$ are calculated by
\begin{align}
	\textbf{V}_i = \frac{\textbf{Q}_{i+1} - \textbf{Q}_i}{\Delta t},  
	\textbf{A}_i = \frac{\textbf{V}_{i+1} - \textbf{V}_i}{\Delta t},
	\textbf{J}_i = \frac{\textbf{A}_{i+1} - \textbf{A}_i}{\Delta t}.
\end{align}

Then the optimization problem is then formulated as:
\begin{align}
	\min_{\textbf{Q}} J = J_s + \lambda_f J_f + \lambda_c J_c + \lambda_d J_d,
\end{align}
where $J_s$ is the smoothness cost defined by
\begin{align}
	J_s = \sum_{i=1}^{N_c-2} ||\textbf{A}_{i}||^2 + \sum_{i=1}^{N_c-3} ||\textbf{J}_{i}||^2,
\end{align}
$J_f$, $J_c$, $J_d$ are penalty terms for the constraints of feasibility, avoiding the static and dynamic obstacles. $\lambda_f$, $\lambda_c$ and $\lambda_d$ are weights for each penalty terms. 

\begin{figure}[t]
	\begin{center}
		\includegraphics[width=1.0\columnwidth]{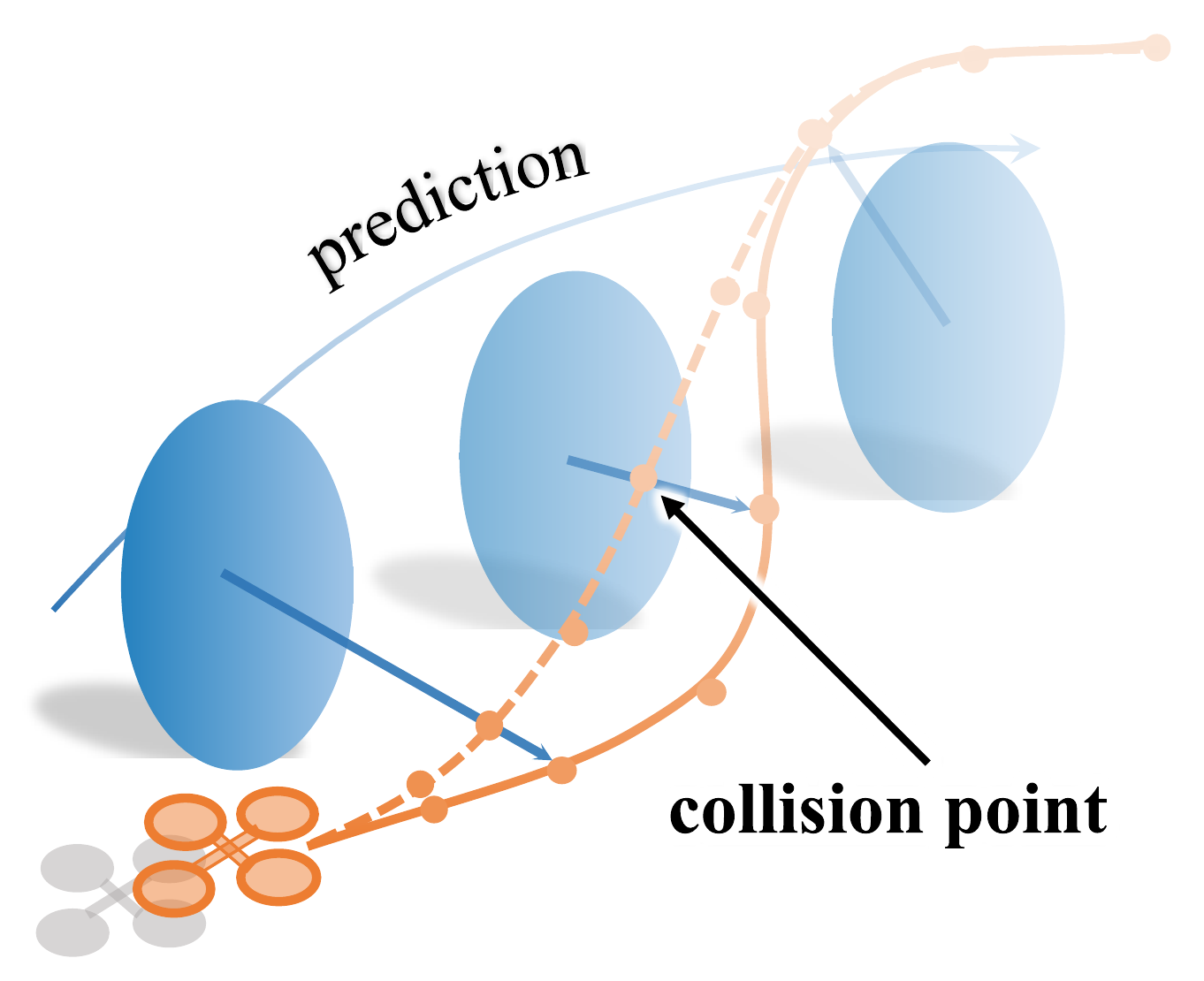}
	\end{center}
	\caption{
		\label{fig:traj}
		An illustration of a initial curve (dotted line) and the corresponding trajectory optimized by dynamic collision penalty (full line). If a control point collides with predicted ellipsoids, the gradient will push it out of moving obstacles.
	}
	\vspace{-0.4cm}
\end{figure}

Thanks to the convex-hull property of B-spline, dynamic feasibility of the whole trajectory can be guaranteed sufficiently if we constrain all the velocity and acceleration control points:
\begin{align}
	J_f = \sum_{i=1}^{N_c-1} g(||\textbf{V}_{i}||^2-v_{m}^2) + \sum_{i=1}^{N_c-1} g(||\textbf{A}_{i}||^2-a_{m}^2), \label{cost function}
\end{align}
where $g(x) = \max\{x,0\}^3$ is a penalty function and $v_{m}, a_{m}$ are the limits on velocity and acceleration.

For static obstacles, we follow the method of our previous work \cite{zhou2020ego}. We firstly search for a collision-free path around static obstacles. Then fixed anchor points and corresponding vectors are generated to determine the static collision penalty and gradients, shown in Fig.\ref{fig:localmin}. The penalty is 
\begin{align}
	J_c = \sum_{i=1}^{N_c} \sum_{j=1}^{N_i}  g\left((\textbf Q_i-\textbf p_{ij} )\cdot \textbf v_{ij} \right),
\end{align}
where $\{\textbf p_{ij}, \textbf v_{ij}\}$ is a pair of anchor point and vector, and $N_i$ is the number of $\{\textbf p_{ij}, \textbf v_{ij}\}$ pairs recorded by the single control point $\textbf Q_i$.

For dynamic obstacles, 
\begin{align}
	J_d = \sum_{i=1}^{N_c} \sum_{j=1}^{N_d}  g\left( 1 - \textbf x_{ij}^T \Theta_j^o \textbf x_{ij} \right),
\end{align}
where $\textbf x_{ij} = \textbf Q_i - \textbf p_j^o - i \cdot \Delta t \cdot \textbf v_j^o$, and $\{\textbf p_j^o, \textbf v_j^o, \Theta_j^o\}$ which represents each ellipsoid is obtained from \ref{section:Dynamic Environment Representation}. A initial curve and its corresponding trajectory optimized by dynamic collision penalty are as shown in Fig.\ref{fig:traj}.

\begin{figure}[t]
	\begin{center}
		\includegraphics[width=1.0\columnwidth]{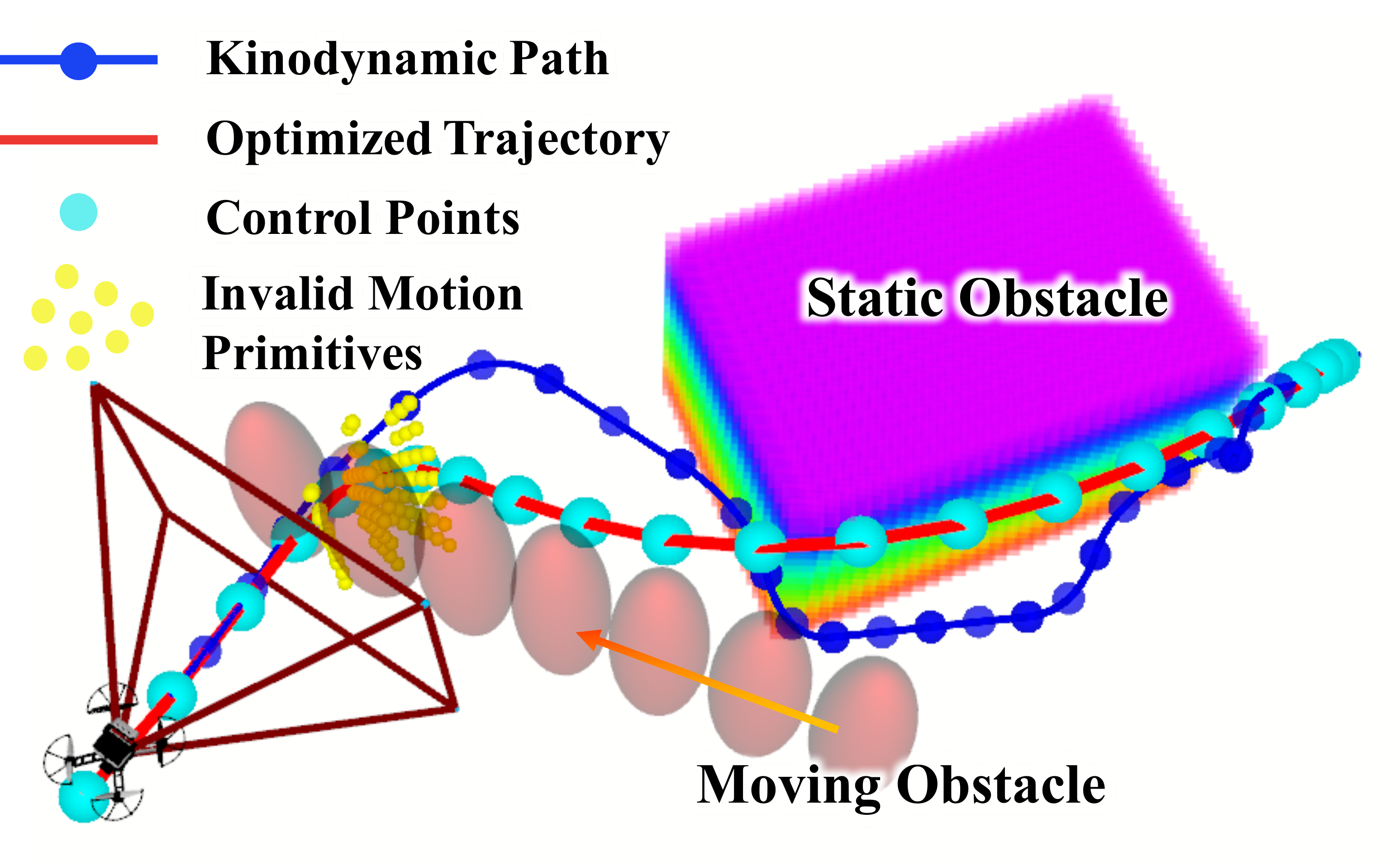}
	\end{center}
	\caption{
		\label{fig:simulation}
		A result of optimization considering both static and dynamic obstacles in simulation. The blue line with blue dots is the path searched by dynamic obstacle-avoiding kinodynamic searching and the red line is the optimized trajectory with control points (cyan points). The yellow points are invalid motion primitives which collide with predicted moving obstacles.
	}
	\vspace{-1.2cm}
\end{figure}

We reparameterize the searching result obtained from \ref{Dynamic Obstacles-avoiding Kinodynamic Searching} to B-spline as the initial guess and utilize the L-BFGS \cite{zhou2020ego} to minimize the cost function (7). A simulated result trajectory of the optimization influenced by both static and dynamic obstacles is shown as Fig.\ref{fig:simulation}.

\section{Experiments and Benchmarks}
\label{sec:experiments}
\subsubsection{Implementation Details}

\begin{figure*}[t]
	\begin{centering}
		\includegraphics[width=2.0\columnwidth]{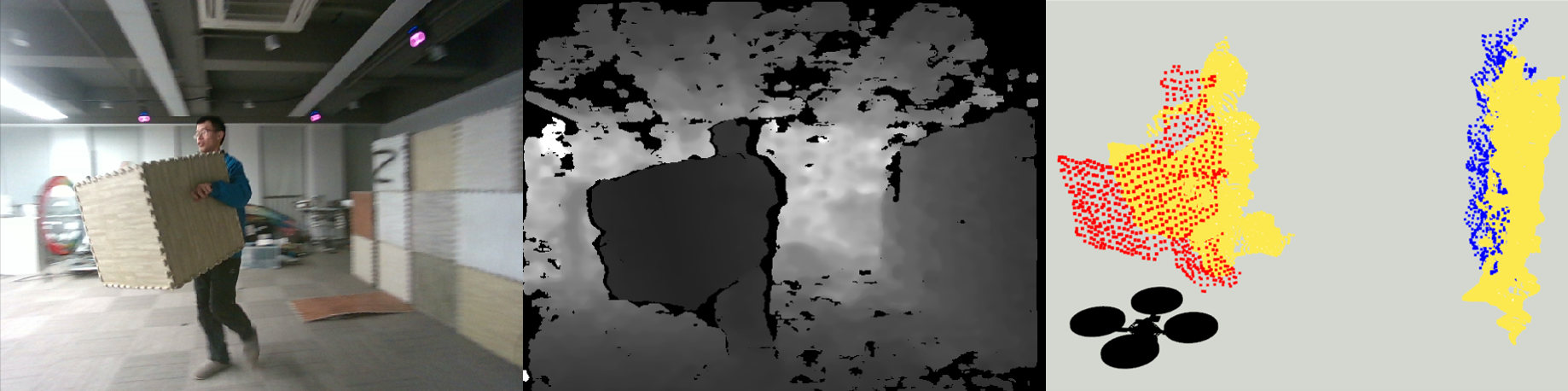}
	\end{centering}
	\caption{
		\label{fig:judge}
		The RGB-D images and the classification result. As the left figure shown, the moving person and left boxes are classified as dynamic and static and marked with red or blue points respectively, while the yellow point cloud is from the reference frame.
	}
	\vspace{0.1cm}
\end{figure*}

In dynamic perception module, we set the time span $\xi = 0.3s$ and the dynamic velocity threshold $v_{nn} = 0.2 m/s$. For a cluster, the absolute and relative threshold of vote for being dynamic are set as $l_{dyn}^{abs} = 100$ and $l_{dyn}^{rel} = 0.8$, and the time horizon is set as $t_h = 0.3s$ to improve the robustness of classification. 

In dynamic planning module, we set the order of B-spline as \emph{p}$_b = 3$. The time span $\Delta t$ of B-spline alters around 100ms, which is determined by the path length searched by our modified kinodynamic A* and the number of control points \emph{N}$_c=20$. We use the same L-BFGS solver as \cite{zhou2020ego} whose complexity is linear on the same relative tolerance. To further enforce safety, a collision checker for both static and dynamic obstacles with enough clearance is performed. The optimization stops only if no collision is detected. In real-world experiments, we adopt a similar quadrotor platform of \cite{gao2020teach} but use Intel RealSense D455 differently, which provides broader FOV.

\subsubsection{Real-World Experiments}

\begin{figure}[t]
	\begin{center}
		\includegraphics[width=1.0\columnwidth]{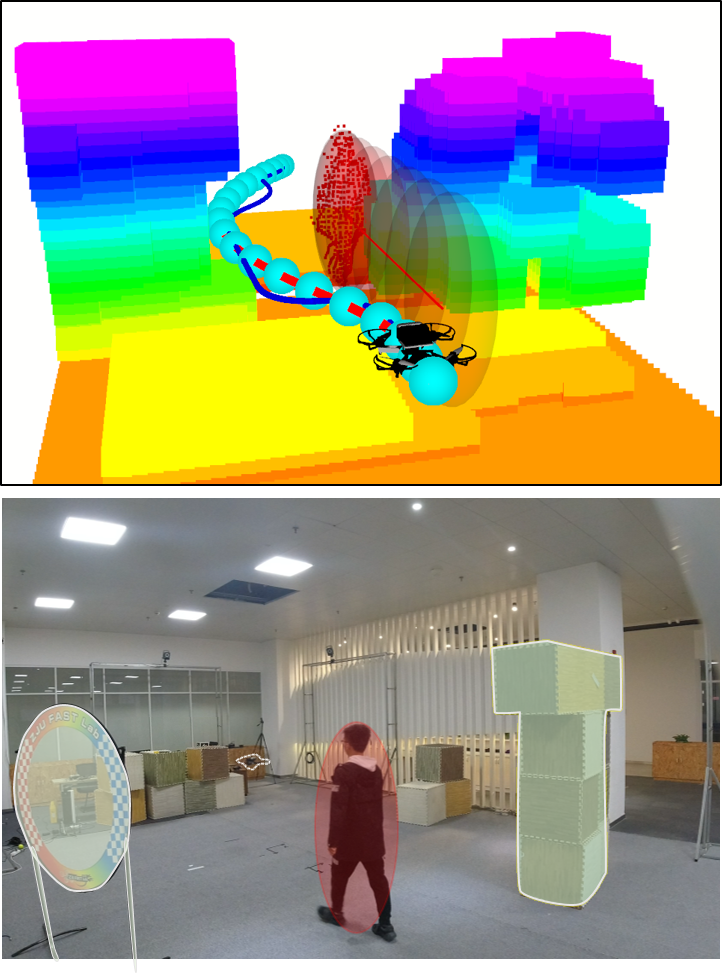}
	\end{center}
	\caption{
		\label{fig:dd} Top: the optimized trajectory between a moving person and static obstacles in narrow space. Bottom: the experiment scenario in which the static obstacles are masked with yellow and the dynamic one is masked with red.
	}
	\vspace{-0.6cm}
\end{figure}

We present several experiments in unknown cluttered environments with limited camera FOV. In the first experiment, the drone flies through static objects, avoiding a man walking toward it simultaneously. As shown in Fig.\ref{fig:dd}, the moving person is modeled as an ellipsoid with velocity, and static obstacles are presented as 3D occupancy grids. The person's speed is around 1.6 m/s and the speed limit of UAV is set as 1.5 m/s. A collision is unavoidable if there is no dynamic perception and prediction. The optimized trajectory also meets the performance in the simulation like Fig. \ref{fig:simulation}.

\begin{figure*}[t]
	\begin{centering}
		\includegraphics[width=2.0\columnwidth]{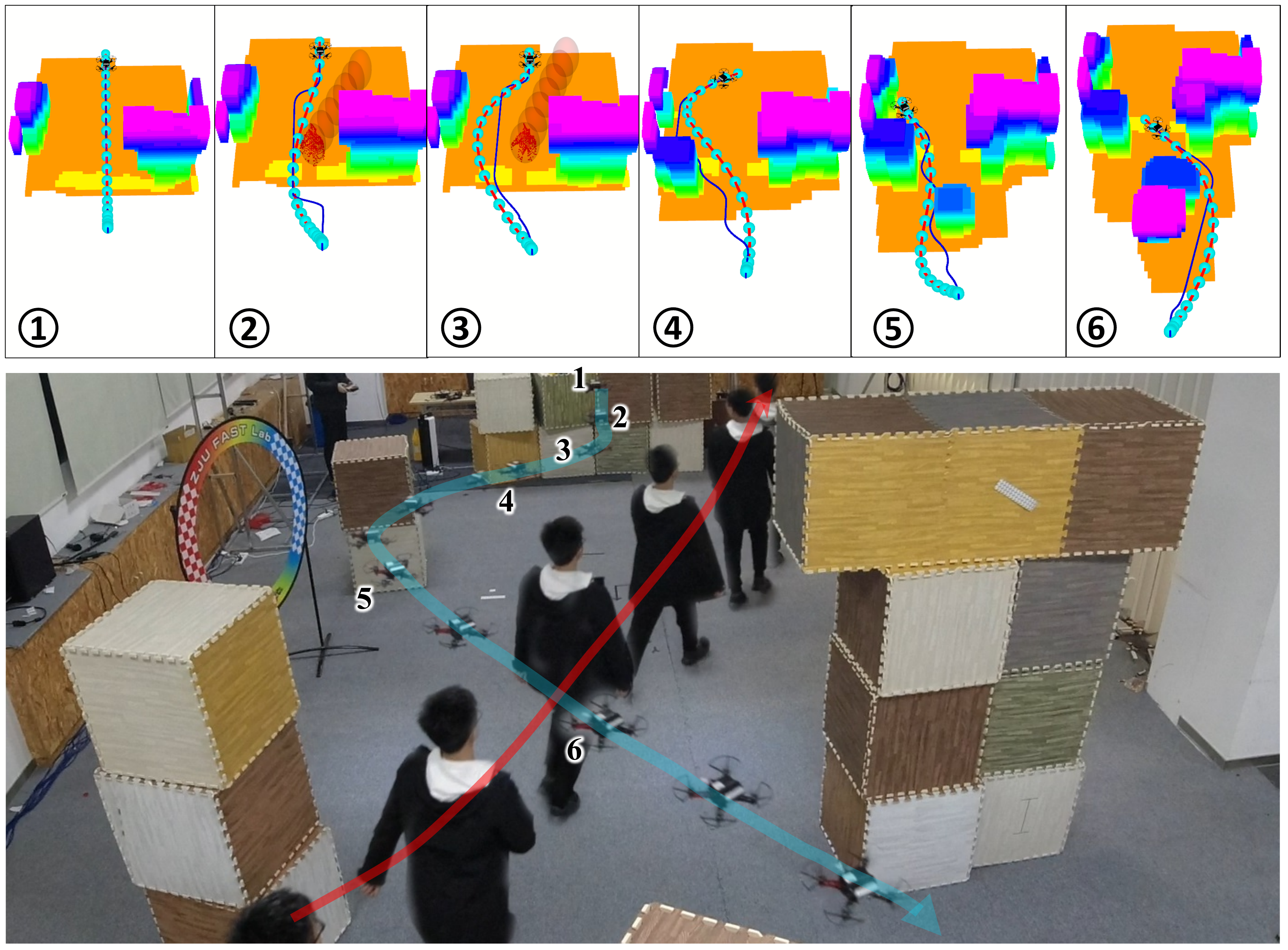}
	\end{centering}
	\caption{
		\label{fig:experiment}
		A sequence of optimized trajectories and corresponding snapshot during the experiment. An UAV is required to fly from a start point to a goal, avoiding an obliquely walking person and a plenty of obstacles. Top: estimated position and velocity of the moving person and optimized trajectory. Bottom: Snapshot of the experiment.
	}
	\vspace{-0.2cm}
\end{figure*}

Another experiment requires the UAV to navigate from a start point to an endpoint while avoiding a man walking across the flight course with higher speed and plenty of static obstacles. Fig.\ref{fig:experiment} shows a series of optimized trajectories in a dynamic environment and a snapshot of the flying UAV and the walking person. With a limited maximum perception range of depth sensor around 4m, the UAV classifies obstacles as dynamic or static, estimates the velocity, and builds a 3D grid map below 30ms, indicating that our system can run efficiently in real-time. The person walks around 2m/s, while the UAV achieves a speed of 2.5m/s in the cluster environment during the experiment. 

In addition, we also conduct plenty of experiments to especially test our proposed  system's robustness. More details are available in the attached video.

\subsubsection{Benchmark Comparisons}

In this section, we both compare the dynamic perception module and dynamic planning module.

Firstly, we compare our dynamic perception module with \cite{lin2020robust}. We carry on both simulated and real-world experiments. In simulated tests, a $40\times10\times3m$ environment is generated with static obstacles and moving agents. In real-world experiments, we record a dataset of which one human walks back and forth in a cluster environment. The experiment's snapshot is shown in Fig.\ref{fig:judge}. We then use both perception methods and compare results with ground truth measurements. The method adopted in \cite{lin2020robust} treats each obstacle as individual ellipsoids without dynamic classification, which results in that the positions and velocities of static obstacles will be frequently misestimated due to the sudden FOV change and occlusion caused by dynamic agents. 

In each comparison above, we calculate the average estimation errors of position and velocity comparing with the ground truth measurements, and the results are shown in Tab.\ref{tab:perception_cmp}. It can be observed that our method could handle more generic environments and perform more accurately.

\begin{table}[t]
	\centering
	\caption{Dynamic Perception Comparison}
	\setlength{\tabcolsep}{1.4mm}
	\renewcommand\arraystretch{1.2}
	{
		\begin{tabular}{|c|c|c|c|c|}
			\hline
			Scenario             & Method       & $error_{pos}$(m) & $error_{vel}$(m/s) \\  \hline
			
			$simulation$ & Ours      &  \bf 0.11        & \bf 0.19           \\ \cline{2-4}
			$experiment$	& Method\cite{lin2020robust} & 0.14       & 0.36    \\ \hline
			$real-world$ & Ours      & \bf 0.18        & \bf 0.29 \\ \cline{2-4}
			$experiment$	& Method\cite{lin2020robust} & 0.31      & 0.57   \\ \hline
	\end{tabular}}
	\label{tab:perception_cmp}
	\vspace{-0.0cm}
\end{table}

\begin{table}[t]
	\centering
	\caption{Dynamic Planning Comparison}
	\setlength{\tabcolsep}{1.4mm}
	\renewcommand\arraystretch{1.2}
	{
		\begin{tabular}{|c|c|c|c|c|c|}
			\hline
			Scenario             & Method       & $t_\text{arrive}$(s) & $l_\text{traj}$(m) & $v_\text{mean}$(m/s) & $t_\text{opt}$(ms) \\  \hline
			
			$N=20,$ & Ours      &  15.75        & 31.1        & 2.07        & \bf5.76             \\ \cline{2-6}
			$v=1m/s$	& Method\cite{zhu2019chance} & 14.7       & 30.0      & 2.04       & 15             \\ \hline
			$N=20,$ & Ours      & 16.2        & 32.1        & 2.02        & \bf 6.34            \\ \cline{2-6}
			$v=2m/s$	& Method\cite{zhu2019chance} & 15.75       & 31.3       & 2.01       &    14          \\ \hline
			$N=50,$  & Ours      & 16.7         & 36.1          & 2.31         & \bf6.27             \\ \cline{2-6}
			$v=1m/s$	& Method\cite{zhu2019chance} & 16.2       & 35.3       & 2.21         & 17            \\
			\hline
	\end{tabular}}
	\label{tab:planner_cmp}
	\vspace{0.05cm}
\end{table}

Afterwards, for dynamic planning, we compare our work with \cite{zhu2019chance}. In \cite{zhu2019chance}, authors assume that locations and velocities of all obstacles are known and do not classify them as dynamic or static. Therefore, we only compare the dynamic planning module in a fully dynamic environment without static obstacles. Each planner runs for twenty times of different obstacle velocities and densities from the same start points to endpoints. In each simulated experiment, obstacles move with constant velocities and reverse when they arrive at the boundary of $20\times20\times3m$ map. The average performance statistics and the computation time are shown in Table.\ref{tab:planner_cmp} 

From Table.\ref{tab:planner_cmp}, we conclude that the performance of our generated trajectory is comparable to \cite{zhu2019chance}. However, our proposed planner needs a much lower computation budget. This is because the time complexity of our method is $O(N_c)$, since one control point only affects nearby segments thanks to the local support property of B-spline, while \cite{zhu2019chance} plans the pose of a quadrotor with nonlinear model predictive control, which considers more constrains. Moreover, our method would not fall into a local minimum, while \cite{zhu2019chance} would.

\section{Conclusion}
\label{sec:conclusion}

In this paper, we introduce a robust vision-based system with onboard sensing for UAV in the dynamic environment and implement it on a computational power-limited quadrotor platform. Firstly we adopt an accurate yet efficient dynamic perception module, which classify obstacles as dynamic or static, obtain necessary state information of moving obstacles, and build a high-quality static occupancy grid map for navigation. Then, our proposed gradient-based planning framework generates local collision-free trajectories without maintaining ESDF by a carefully designed optimization in less than 7ms. Sufficient real-world experiments and benchmark comparisons validate that our system is effective, robust, and highly lightweight in unknown cluster environment.

\addtolength{\textheight}{0.7cm} 

\bibliography{RAL2021}
\end{document}